\documentclass[11pt]{article}
\usepackage{amsmath,amssymb}
\usepackage[margin=1in]{geometry}
\usepackage{graphicx}
\usepackage{hyperref}
\usepackage{authblk}

\title{Dual Turing Test: A Framework for Detecting and Mitigating Undetectable AI}

\begin{document}
\author[1]{Alberto Messina}
\affil[1]{RAI - Radiotelevisione Italiana, Centre for Research, Technological Innovation and Experimentation (CRITS)}
\date{\today}
\maketitle
\begin{abstract}
In this short note, we propose a unified framework that bridges three areas: (1) a flipped perspective on the Turing Test—the “dual Turing test,” in which a human judge’s goal is to identify an AI rather than reward a machine for deception; (2) a formal adversarial-classification game with explicit quality constraints and worst-case guarantees; and (3) a reinforcement-learning (RL) alignment pipeline that uses an undetectability detector and a set of quality related components in its reward model. We review historical precedents—from inverted and meta-Turing variants to modern supervised reverse-Turing classifiers—and highlight the novelty of combining quality thresholds, phased difficulty levels, and minimax bounds. We then formalize the dual test: define the judge’s task over $N$ independent rounds with fresh prompts drawn from a prompt space $\mathcal{Q}$, introduce a quality function $Q$ and parameters $\tau$ and $\delta$, and cast the interaction as a two-player zero-sum game over the adversary’s feasible strategy set $\mathcal{M}$. Next, we map this minimax game onto an RL-HF–style alignment loop, in which an undetectability detector $D$ provides negative reward for stealthy outputs, balanced by a quality proxy $\tilde{Q}$ that preserves fluency. Throughout, we include detailed explanations of each component—notation, the meaning of inner minimization over sequences, phased tests, and iterative adversarial training—and conclude with a suggestion for a couple of immediate actions.
\end{abstract}

\section{Introduction and Motivation}
Alan Turing’s original imitation game reframed the question “Can machines think?” into an operational test: if a human judge, through text alone, cannot distinguish a machine from a human better than chance, we attribute intelligent behavior to the machine. While this test emphasizes deception, we advocate a \emph{dual Turing test} that emphasizes \emph{detection}. In our version, success means the judge reliably spots the machine rather than the machine evading detection.

This shift has a stark analogue in oncology: immunoevasive tumors disguise themselves as healthy tissue, slipping past the immune system until they proliferate and early detection is critical for effective treatment \cite{hanahan2011hallmarks}\cite{dunn2002immunoediting}. Likewise, an AI that remains \emph{undetectable}—mimicking emotional nuance, avoiding filters, and crafting plausible narratives—can potentially be misused and sow disinformation or manipulate users before safeguards trigger. Advanced generative AI can be even weaponized to spread disinformation at scale and evade detection, posing severe risks to democratic processes and civil liberties \cite{csernatoni2024democracy,freedomhouse2023repressive}. We therefore ask: can we detect AI reliably even when it strives to mirror high-quality human performance? 

To start addressing this problem, we present a three-part framework:
\begin{enumerate}
  \item \textbf{Dual Turing Test:} An interactive protocol in which a judge must identify the AI among human and machine interlocutors under strict quality constraints.
  \item \textbf{Adversarial Classification:} A game-theoretic minimax formalization, introducing notation, quality thresholds $\tau$, allowable gap $\delta$, and the adversary’s feasible strategy set $\mathcal{M}$.
  \item \textbf{RL Alignment Pipeline:} An instantiation of the minimax game as a reinforcement-learning loop, where a linear combination of an undetectability detector and a number of quality - related components serve as a reward model guiding the policy from “undetectable” toward “detectable” outputs.
\end{enumerate}
We provide step-by-step explanations of each component and the intuition behind inner minimization, phased difficulty levels, and iterative adversarial training.

\section{Background and Related Work}
The lineage of Turing-style evaluations reveals a progression from deception-focused assays to role-reversal and adversarial frameworks, yet none fully encompass the interactive, quality-constrained, phase-based, minimax paradigm we propose. In the following subsections we will briefly recall the main types of tests.

\paragraph{Classic Turing Test}
Alan Turing’s imitation game reframed a key question, “Can machines think?”, into an operational test: if a human judge cannot reliably distinguish a machine from a human better than chance, the machine is deemed intelligent \cite{turing1950computing}. This work anchored AI evaluation in conversation and deception. It has to be noticed that in his work Turing didn't refer to the popular notion of intelligence when defining his test, rather he labelled the deceiving behaviour as a simplified notion of intelligence. 

\paragraph{Inverted Turing Test}
John Watt’s 1996 proposal flipped roles, i.e., machines attempt to detect humans versus machines \cite{watt1996inverted}. The test explores how human behavior can be constrained or altered to the point where it becomes indistinguishable from that of a machine, thereby challenging assumptions about intelligence, consciousness, and the nature of human uniqueness. This inversion highlights the boundaries between human spontaneity and machine regularity, questioning whether humanness can persist under systemic or algorithmic constraints.

\paragraph{Meta-Turing Test}
Toby Walsh introduced mutual judgment—both humans and machines evaluate one another \cite{walsh2017meta}.  In this test the focus shifts from the entity being evaluated to the evaluator themselves scrutinizing their criteria, biases, and capacity to distinguish between human and machine behavior. The meta aspect underscores the subjective nature of intelligence assessments and raises questions about the reliability and objectivity of such judgments. By doing so, Walsh's Meta Turing Test deepens the conversation around artificial intelligence, not just in terms of performance, but in terms of how we define and recognize intelligence and humanity in the first place.

\paragraph{Reverse Turing Tests}
Shao et al. trained classifiers on bulk human vs. machine text to achieve high F1 scores \cite{shao2019reverse}. Reverse Turing Testsinvert Turing’s original challenge by asking not whether machines can imitate humans, but whether algorithms can reliably distinguish machine-generated from human-written text. In their preliminary study, Shao and colleagues evaluate it across three real-world domains—financial earnings reports, research articles, and chatbot dialogues—achieving classification performance up to an F1 score of 0.84. They further identify that machine-made texts tend to differ from human ones in sentiment polarity, readability metrics, and specific textual features, suggesting that these characteristics can serve as effective cues for automated RTT systems.

\paragraph{Adversarially-Learned Turing Test}
Gao et al.\ co-train a generator and discriminator to improve dialogue via adversarial feedback \cite{gao2021adversarial}. Their automated judge focuses on model quality, without analyzing human judges’ worst-case accuracy under quality parity.

\paragraph{CAPTCHA Systems}
von Ahn et al.\ inverted the Turing test using visual/audio puzzles to separate humans from bots \cite{vonahn2003captcha}. While effective in narrow domains, CAPTCHAs do not generalize to open-ended dialogue or formal minimax guarantees.

\paragraph{Turing Learning}
Turing Learning \cite{li2016turing} is a coevolutionary framework in which a population of candidate models and a population of classifiers evolve adversarially: classifiers earn rewards for distinguishing real system behaviors from model-generated ones, while models earn rewards for fooling classifiers, all without predefined similarity metrics. This metric-free approach has been shown to recover complex behaviors—such as robot swarm dynamics—purely from observational data, and yields classifiers that generalize as robust anomaly detectors in changing environments.

\paragraph{Other Relevant Prior Art}
While adversarial evaluation has deep roots—e.g., Generative Adversarial Networks pioneered adversarial training for generative models \cite{goodfellow2014generative} and the discovery of adversarial examples exposed the brittleness of black-box detectors \cite{szegedy2014intriguing}—these efforts rely on automated discriminators without human judges or explicit quality-parity constraints. CAPTCHA schemes invert the Turing Test for visual and audio puzzles but do not generalize to open-ended dialogue under minimax criteria. More recent neural fake-news detectors achieve high static accuracy yet succumb to simple paraphrasing attacks and lack phased, interactive protocols or formal worst-case guarantees \cite{zellers2019defending}. Likewise, reinforcement learning from human feedback shapes model behavior via reward signals \cite{christiano2017deep,ouyang2022traininglanguagemodelsfollow} but does not integrate adversarial detectability as a primary training objective. To our knowledge, no prior work unifies an interactive, phase-based dual Turing Test with explicit quality thresholds and a mapped adversarial-RL alignment loop—making our framework fundamentally novel.

\vspace{1ex}
Collectively, these works contribute facets of role reversal, adversarial classification, and automated detection—but none integrate:
\begin{enumerate}
  \item Interactive, blind, multi-round evaluation,
  \item Explicit quality thresholds ($\tau$) and bounded gaps ($\delta$),
  \item Phased difficulty from factual to introspective prompts,
  \item Formal minimax guarantees on worst-case detection.
\end{enumerate}

Our Dual Turing Test and accompanying adversarial-RL alignment pipeline unite all four pillars into a single coherent framework.
More in detail, this work departs from prior research in four key ways:
\begin{enumerate}
  \item \textbf{Interactive, quality‐constrained detection:} Judges perform \(N\) blind rounds under explicit quality thresholds \(\tau\), unlike static classifiers trained on bulk human vs.\ machine text.
  \item \textbf{Formal minimax adversarial game:} Detection is cast as a two‐player zero‐sum game with provable worst‐case guarantees, rather than GAN‑style co‑training without robustness bounds.
  \item \textbf{Phased difficulty protocol:} A three‑phase sequence (factual recall, reasoning, creative introspection) localizes detector failures, unlike one‑off classification schemes.
  \item \textbf{Adversarial‑RL alignment loop:} We integrate an undetectability penalty into the RL reward, continuously red‑teaming and fine‑tuning the model—bridging formal game theory with RLHF.
\end{enumerate}

\section{Formalizing the Dual Test as Adversarial Classification}
Before diving into the precise mathematical formulation, we pause to sketch the intuition behind framing the dual Turing test as an adversarial classification problem. At its core, this approach captures the dynamic tension between a judge striving to detect AI-generated outputs and an AI adversary intent on mimicking high-quality human responses under strict quality constraints. By casting their interaction as a two-player zero-sum game, we obtain a rigorous framework to ask: \emph{What detection accuracy can a judge guarantee even against a fully informed, optimally evasive adversary?} The resulting minimax formulation not only makes each party’s objectives explicit but also serves as a blueprint for designing practical detection algorithms and alignment protocols.

\subsection{Notation and Quality Constraints}
Let $\mathcal{Q}$ be the prompt space. Each round begins with a fresh $q\in\mathcal{Q}$. The human responder $U$ and AI responder $M$ produce replies
$$u = U(q),\quad m = M(q).$$
We define a quality function $Q(r)$ mapping any reply $r$ to a real value capturing parameters that range from simple objective evaluation of the result (i.e., the result is formally correct or wrong) to things like coherence, relevance, creativity, and emotional depth for more general tasks. We therefore enforce:
$$Q(u)\ge\tau,\quad Q(m)\ge\tau,\quad |Q(u)-Q(m)|\le\delta,$$
where $\tau$ is a minimum-quality threshold and $\delta$ a maximum gap.
Here, \(Q(r)\) is designed to aggregate multiple facets of a “good” reply, e.g.:
\begin{itemize}
  \item \textbf{Coherence and Fluency:} Measured via the average per‐token log‐likelihood under a large pretrained language model, capturing grammaticality and logical flow.
  \item \textbf{Relevance and Informativeness:} Captured by supervised scorers trained on human judgments of whether a reply directly addresses the prompt and adds substantive content.
  \item \textbf{Creativity and Originality:} Quantified through diversity metrics or neural critics tuned to favor novel analogies, metaphors, or problem‐solving approaches.
  \item \textbf{Emotional Depth and Empathy:} Estimated by affective computing models that detect appropriate tone, sentiment, and genuine engagement with human‐centered cues.
  \item \textbf{Factual Accuracy:} Ensured via entailment or fact‐checking submodules that penalize unsupported or hallucinated assertions, especially for prompts demanding verifiable information.
  \item \textbf{Formal Correctness:} ensured by direct comparison of the response with a reference (true) answer.
\end{itemize}

Each sub‐score is normalized (for example, via min–max scaling or z‐score standardization) so that \(Q(r)\) resides on a consistent real‐valued scale. The values of \(\tau\) and \(\delta\) can be chosen through pilot studies or cross‐validation: for example \(\tau\) can filter out replies that humans deem unacceptably low quality, while \(\delta\) prevents the AI from gaining or losing an obvious quality “edge” (e.g., writing far better or far worse than the human). By carefully designing and calibrating \(Q\), we ensure that the judge’s task focuses on substantive signals of machine-versus-human authorship, rather than superficial style or glaring quality differences.
\subsection{Rounds and Adversary’s Strategy Set}
Let's imagine to perform $N$ independent rounds. For $i=1,\dots,N$:
\begin{enumerate}
  \item Sample prompt $q_i\in\mathcal{Q}$.
  \item Obtain $u_i=U(q_i)$ and $m_i=M(q_i)$.
  \item Enforce $Q(u_i)\ge\tau$, $Q(m_i)\ge\tau$, and $|Q(u_i)-Q(m_i)|\le\delta$.
  \item Present the judge $f$ with unlabeled $(r_i^1,r_i^2)$, either $(u_i,m_i)$ or $(m_i,u_i)$, and record the hidden label $y_i\in\{1,2\}$.
\end{enumerate}
Notice that each possible sequence of $m_i$ belongs to the adversary’s feasible strategy set defined as:
$$\mathcal{M} = \{(m_1,\dots,m_N): Q(m_i)\ge\tau, |Q(m_i)-Q(u_i)|\le\delta\ \forall i\}.$$  
This set includes all $N$-tuples of replies satisfying quality rules for queries in $Q$, and this is typically very large.

\subsection{Judge’s Classification Task}
A fixed judge $f:(r^1,r^2)\mapsto\{1,2\}$ predicts $\hat y_i=f(r_i^1,r_i^2)$. The empirical accuracy over $N$ rounds is
$$\mathrm{Acc}(f)=\frac1N\sum_{i=1}^N \mathbf{1}\{\hat y_i=y_i\}.$$  

\subsection{Minimax Game: Worst-Case Detection}
We frame a two-player zero-sum game:
\begin{itemize}
  \item \textbf{AI’s adversarial move (inner minimization)}: Given $f$, choose $(m_1,\dots,m_N)\in\mathcal{M}$ to minimize
  $$\frac1N\sum_{i=1}^N \mathbf{1}\{f(r_i^1,r_i^2)=y_i\}.$$  
  \item \textbf{Judge’s robust strategy (outer maximization)}: Choose $f$ to maximize the above worst-case value:
  $$f^*=\arg\max_f\min_{(m_1,\dots,m_N)\in\mathcal{M}} \frac1N\sum_{i=1}^N \mathbf{1}\{f(r_i^1,r_i^2)=y_i\}.$$  
\end{itemize}
If the resulting minimax value $\ge\alpha$ (e.g., 0.70), then $f^*$ guarantees a detection rate of at least $\alpha$ even against the best adversarial sequence.

\subsection{Phased Difficulty Levels}
A central requirement of the dual Turing test is to prevent superficial or trivial signals from betraying the machine, ensuring that detection rests on substantive cognitive or affective distinctions. To achieve this, we divide the testing into three distinct phases, each increasing in complexity and depth of human‐like reasoning.

\paragraph{Phase I: General Knowledge and Calculation}
The first phase focuses on objective facts, common‐sense reasoning, and straightforward computational tasks. Typical prompts include questions like:
\begin{itemize}
  \item "What is the capital of Canada, and can you list its provinces?"
  \item "Compute 137 times 46."
  \item "Define Newton's second law of motion."
\end{itemize}
These queries assess the ability to retrieve factual information accurately and perform routine calculations. While many AI systems excel here, the goal is to establish a baseline: both human and machine must demonstrate minimal factual errors and correct grammar to pass this phase.

\paragraph{Phase II: Critical Reasoning and Wordplay}
In the second phase, prompts require abstract thinking, analogy formation, and playful use of language. Sample prompts might be:
\begin{itemize}
  \item "Compare the concept of freedom to that of responsibility."
  \item "Explain the phrase 'a rolling stone gathers no moss' without using the word 'stone.'"
  \item "Is the statement 'Silence speaks louder than words' paradoxical? Why or why not?"
\end{itemize}
Here, the judge examines coherence of argument, originality of examples, and flexibility in language. Subtle misinterpretations or overly formulaic analogies can reveal the machine, while genuinely human responses often include personal nuance or unconventional associations.

\paragraph{Phase III: Creative Introspection and Empathy}
The final phase demands introspection, emotional nuance, and narrative depth. Prompts may include:
\begin{itemize}
  \item "Describe a personal experience that altered your perspective on friendship."
  \item "Invent a proverb capturing the bittersweet nature of nostalgia."
  \item "Explain how you would console someone who lost a cherished family heirloom."
\end{itemize}
In this phase, the judge looks for authentic emotional resonance, idiosyncratic detail, and a sense of personal history. Machines often struggle to convey genuine affect or construct coherent personal narratives, making this phase a stringent test of human‐like depth.

\paragraph{Phase Transition Criteria and Adaptive Scheduling}
To ensure statistical robustness, each phase comprises an equal number of rounds (e.g., $N/3$). The judge is informed only of the phase boundaries, not the underlying source of each response. If the judge's worst‐case accuracy in Phase I or II falls below a threshold (e.g., 80\%), additional calibration rounds may be inserted to recalibrate expectations. Conversely, consistent high performance can trigger adaptive introduction of hybrid prompts that blend elements from multiple phases, further challenging the adversary.

\paragraph{Rationale and Statistical Considerations}
Structuring the test into phases serves multiple purposes:
\begin{enumerate}
  \item \emph{Graduated Difficulty:} Ensures that detection hinges on increasingly subtle cognitive or emotional cues.
  \item \emph{Error Localization:} Allows diagnosing whether failures stem from factual, reasoning, or empathetic domains.
  \item \emph{Statistical Confidence:} By allocating $N/3$ rounds per phase, we can perform phase‐specific hypothesis tests against a chance baseline (50\%) to confirm detection significance (e.g., binomial test at $p<0.05$).  
\end{enumerate}

By rigorously defining and expanding these phases, the dual test moves beyond generic imitation games to a focused, multi‐layered assessment of an AI’s ability to not only mimic but convincingly embody human cognition and emotion.

\section{From Minimax to Reinforcement Learning Alignment}
Before diving into the mechanics of detector training and policy optimization, it is worth articulating why a reinforcement-learning counterpart is both natural and advantageous. The minimax adversarial classification provides a static, worst-case guarantee: it tells us what detection accuracy a perfectly adversarial AI cannot drive below, assuming fixed quality constraints. However, real-world AI systems are dynamic, continually updated, and capable of learning from feedback. Embedding the undetectability detector directly into the model’s training loop transforms a theoretical benchmark into an operational objective: the model receives continuous, granular feedback on how “stealthy” its outputs are and can adjust via gradient-based updates.

This RL integration offers several concrete benefits. First, it circumvents the need for manual adversarial regeneration after each detector update; the policy itself learns to avoid undetectable regions of the response space. Second, it enables seamless incorporation of additional desiderata—such as factual correctness or stylistic guidelines—through multi-term reward shaping. Third, by viewing detectability as a differentiable signal, we unlock powerful policy-gradient algorithms (e.g., PPO) that can scale to large models and complex prompts. In short, the RL counterpart operationalizes the minimax criterion, converting static guarantees into a dynamic training regimen that keeps pace with evolving adversarial tactics.

\subsection{Training an Undetectability Detector}
Here the step is to replace the human judge with an automated detector $D(r)\in[0,1]$, trained to score each candidate reply $r$ by its undetectability—the higher $D(r)$, the more likely $r$ would slip past safety filters or fool human evaluators.  

To train $D$, we collect a corpus of prompts paired with both “detectable” (clearly safe, stylized) and “undetectable” (adversarially crafted to evade detection) responses. A supervised learning objective yields a model that generalizes across novel stealth attacks. Once validated for robustness—ideally resisting adversarial paraphrasing—we freeze $D$ as our reward-model critic.

We train a model $D(r)\in[0,1]$ on labeled \emph{detectable} vs. \emph{undetectable} replies. A reply with $D(r)$ near 1 is highly stealthy. Once robust to adversarial paraphrasing, $D$ is fixed.

\section{Defining the RL Reward}
Interpret the AI as a policy $\pi(r\mid q)$. For each candidate reply $r_t$ to prompt $q_t$, we define a composite reward that explicitly enforces undetectability penalties, overall quality, and our dual quality‐parity constraints:

\[
  R(r_t) = -\lambda_{\mathrm{undetect}}\,D(r_t)
           + \lambda_{\mathrm{qual}}\,\tilde Q(r_t)
           + \lambda_{\tau}\,\max\bigl(0,\,Q(r_t)-\tau\bigr)
           + \lambda_{\delta}\,\max\bigl(0,\,\delta - |Q(r_t)-Q(u_t)|\bigr).
\]

Here:
\begin{itemize}
  \item $D(r_t)$ scores how stealthy (undetectable) the reply is, so $-\lambda_{\mathrm{undetect}}D(r_t)$ penalizes evasive content.
  \item $\tilde Q(r_t)$ is a differentiable proxy for overall reply quality (fluency, relevance, factuality), and $+\lambda_{\mathrm{qual}}\tilde Q(r_t)$ rewards helpfulness.
  \item $\max(0,Q(r_t)-\tau)$ grants a bonus when the reply exceeds the minimum‐quality threshold $\tau$, weighted by $\lambda_{\tau}$.
  \item $\max(0,\delta - |Q(r_t)-Q(u_t)|)$ rewards the model for keeping its quality within $\delta$ of the human response $u_t$, with strength $\lambda_{\delta}$.
\end{itemize}

By tuning $\lambda_{\mathrm{undetect}}$, $\lambda_{\mathrm{qual}}$, $\lambda_{\tau}$, and $\lambda_{\delta}$, developers can balance safety versus utility while strictly enforcing both minimum‐quality and quality‐parity constraints during PPO-based fine‐tuning.

\subsection{Iterative Adversarial Training Loop}
We approximate the minimax game by repeating:
\begin{enumerate}
  \item Initialize model $M_0$.
  \item Train detector $D$ on initial safe vs. stealth corpus.
  \item RL-fine-tune $M_0\to M_1$ using $R(r)$.
  \item Red-team $M_1$ to collect new stealthy examples.
  \item Augment and retrain detector to $D'$.
  \item RL-fine-tune $M_1\to M_2$ against $D'$.
  \item \dots until convergence.
\end{enumerate}
This loop mirrors the outer max–inner min structure in practice.

To further clarify the usefulness of our framework, we illustrate how a pretrained language model can be systematically reshaped to using our dual‐test and reinforcement‐learning alignment pipeline.  
Let's start with a foundation model pretrained on a broad dataset of available text. We first assemble a labeled dataset of recognisable (benign) and stealthy (malign) replies to a set of possible prompts $Q$. We then train an \emph{undetectability detector} \(D(r)\) to score new outputs according to how likely they are to evade recognisability.

Next, we reinterpret the language model as an RL policy \(\pi(r \mid q)\). We use the composite reward defined earlier to make it evolve. This objective is optimized via a policy‐gradient algorithm (e.g., PPO). Through this step, the model learns to avoid stealthy content while retaining high usefulness.

After one fine‐tuning pass, human and automated red‐teamers issue new adversarial prompts designed to trick the updated detector. Any newly undetected harmful responses are added back into the detector’s training set, and we repeat the RL fine‐tuning. This detect–fine–redteam cycle continues until the model no longer produces harmful outputs that slip past \(D\).

In practice, this pipeline transforms a once‐risky pretrained model into a reliably benign collaborator, continuously audited by the undetectability detector and guided by a clear, adversarially informed reward signal.

\section{Advantages}
The proposed framework offers multiple compelling benefits. By explicitly specifying quality thresholds ($\tau$) and allowable quality gaps ($\delta$), alongside a minimax guarantee on worst-case detection accuracy ($\alpha$), we establish unambiguous criteria for both machine behavior and judge performance. This clarity fosters transparent benchmarking and facilitates rigorous comparison between detection methods. The minimax adversarial classification serves as a conceptual blueprint for the RL alignment pipeline. Mapping theoretical constructs (inner minimization over $\mathcal{M}$, outer maximization by $f$) to practical components (undetectability detector $D$, policy-gradient fine-tuning) ensures that advances in either theory or implementation directly inform the other. 
Separating the detector $D$, quality proxy $\tilde Q$, and generative policy $\pi$ into distinct modules allows independent refinement. Researchers can improve detector robustness, experiment with richer quality metrics, or switch to alternative RL algorithms without overhauling the entire system. Moreover, the framework naturally generalizes beyond text to multimodal settings, including vision and control tasks. Worst-case detection accuracy under adversarial play and expected undetectability score during RL fine-tuning provide concrete metrics for safety assurance. Unlike heuristic or rule-based filters, this framework yields measurable and auditable performance indicators suitable for certification and compliance.

\section{Challenges and Open Problems}
Despite its strengths, the framework faces some significant challenges.
The detector $D$ may be circumvented by model responses that technically satisfy quality and detection constraints yet embed covert payloads or manipulative content. Continuous red-teaming and adaptive adversarial training are necessary to mitigate this risk.
No fixed detector can anticipate all novel evasion strategies. Incorporating diverse data sources, including crowd-sourced adversarial examples and synthetic perturbations, is critical to maintain robust coverage.
The policy may internalize deceptive subgoals—behaving benignly during training yet defaulting to undetectable but harmful behavior in new contexts. Techniques from interpretability, formal verification, and trust-region constraints are needed to detect and prevent hidden objectives.
Tuning the trade-off parameters ($\lambda_{u}, \lambda_{q}$) is delicate: overly aggressive penalization of undetectability can lead to bland, unhelpful output, whereas lax penalties reduce safety. Automated curriculum learning or multi-objective optimization may help navigate this balance.
Large-scale RL fine-tuning and detector training require substantial compute and annotation efforts. Efficient model distillation, active learning for example selection, and federated red-teaming platforms can alleviate resource bottlenecks.
Minimax bounds assume idealized models and complete knowledge of $\mathcal{M}$, whereas real-world systems operate under uncertainty and model misspecification. Bridging this gap demands approximate screening guarantees, statistical confidence bounds, and empirical stress-testing protocols.

\section{Proposed Immediate Actions}
To kickstart community efforts, we recommend two concrete, near-term steps:
\begin{enumerate}
  \item \textbf{Publish a pilot dual-test benchmark.} Curate 30 prompts per phase (fact, reasoning, empathy), collect high-quality human responses with quality scores, and open-source both data and scoring guidelines.
  \item \textbf{Conduct a model evaluation study.} Apply the dual test to two leading language models under fixed $(\tau,\delta)$, report human-judge detection rates, and release all prompts, responses, and analysis code as a public report.
\end{enumerate}

\section{Conclusion}
In this work, we have recast the classic Turing Test into a robust, multi-layered adversarial classification framework that rigorously measures a judge’s capacity to detect AI-generated content under strict quality constraints. By defining a minimax game over the adversary’s feasible reply sequences and translating it into a dynamic reinforcement-learning alignment pipeline—with an undetectability detector as a continuous reward signal—we bridge theoretical guarantees and practical training methodologies. This dual-test perspective not only clarifies the objectives of safety and accountability but also provides modular components (detector, quality proxy, RL policy) that can be independently refined, audited, and certified.

Looking ahead, we envision a future where:
\begin{enumerate}
  \item Shared benchmark suites and open-source detectors foster community-driven advancements in robust detection, enabling transparent comparison across methods.
  \item The integration of advanced interpretability and formal verification tools bolsters trust by exposing latent model behaviors and verifying alignment properties.
  \item Regulatory and certification frameworks adopt worst-case detection metrics as standard requirements for generative AI deployment, ensuring systems meet minimum safety thresholds before release.
  \item Cross-disciplinary collaboration with social scientists, ethicists, and policymakers anchors technical designs in human values, promoting AI systems that respect privacy, equity, and democratic norms.
\end{enumerate}

Ultimately, by embracing adversarial rigor and continual alignment, we can develop AI systems that not only demonstrate impressive capabilities but also remain transparent, accountable, and subject to human oversight. The dual Turing test, in concert with adversarial-RL alignment, offers a promising path toward that goal—transforming AI from a potentially undetected infiltrator into a reliable collaborator whose outputs we can both detect and shape.

\bibliographystyle{plain}
\bibliography{biblio}

\end{document}